# Deep Learning for Material recognition: most recent advances and open challenges

Alain Trémeau, Sixiang Xu, Damien Muselet,

## Introduction

Computational materials design is a recent emerging research area aiming at designing, simulating, predicting innovative materials. Among recent innovative applications some are discussed in [Ana Serrano 2016, Yue Liu 2017, Hubert Lin 2018]). The latest advancements in machine learning domain have highly revolutionized computational and data-minded methodologies used for materials design innovation and materials discovery and optimization [Yue Liu 2017, Mosavi A. 2017, Wencong Lu 2017, J. E. Gubernatis 2018]. Virtual design and simulation of innovative materials, as well as recognition of materials from their appearance, require modeling the fundamental properties of materials and of their appearance. This paper reviews the state of the art machine learning methods that have brought advances in the domain of materials recognition from their appearance.

Classifying materials from an image is a challenging tasks for humans and also for computer systems. Few recent papers show the potential of Convolutional Neural Network (CNN) techniques and Support Vector Machine (SVM) classifiers to train architectures to extract features in order to achieve outstanding classification of a diversity of materials. Some years ago, the potential of Convolutional Neural Networks (CNN) on the classification of fine-grained images, such as materials, was controversial. However few papers recently demonstrated the performance of CNNs for material recognition (e.g. see [Li Liu 2019, Raquel Bello-Cerezo 2019]). Thus, [Raquel Bello-Cerezo 2019] demonstrated the superiority of deep networks and of off-the-shelf CNN-based features, particularly with non-stationary spatial patterns, such as textures, and in the presence of multiple changes in the acquisition conditions, against traditional, hand-crafted descriptors. In [Anca Sticlaru 2017] the author evaluated and compared a selection of CNN architectures on various widely used material databases and achieved up to ~92.5% mean average precision using transfer learning on MINC 2500 material database.

In order to improve the efficiency of classifiers, several training sets optimization strategies could be investigated:
- create new databases of images having domain-specific material properties (e.g. databases of translucent or glossiness materials);
- create bigger databases of images with several representatives per material category acquired under a multitude of different viewing and illumination conditions, acquired under controlled and uncontrolled settings (e.g. datasets generated from an internet image database such as Flickr Material Database (FMD) [Sharan 2014]);

- combine different datasets, as in [Grigorios Kalliatakis 2017], or trains a network from one generalist dataset and fine-tunes the network using another more specialized dataset, as in [P. Wieschollek 2016]. The diversity of the datasets is also useful to avoid bias occurring between the training and the test sets;
- use "data augmentation" techniques (e.g. using synthesize appearance, as in [Michael Weinmann 2014, Maxim Maximov 2018], using active learning to grow small datasets, as in [J. E. Gubernatis 2018];
- create large-scale databases of images to better train deep neural networks, such as ImageNet [O. Russakovsky 2014];
- learn view-independent appearance features (or shape-independent appearance features as in [Manuel Lagunas 2019]), learn context-independent appearance features;

In Sections 1 and 2, we will survey the state of the art to show which strategies have already been investigated in the literature.

On the other side, in order to improve the accuracy and robustness of classifiers several data mining strategies have been proposed in the literature:

- compensate with CNN the imbalance between classes (e.g. see [Mateusz Buda 2018]) that often happens between material categories, such as MINC (see Table 1).
- use deeper deep neural networks,
- use transfer learning models, such as in [M. Cimpoi 2015, P. Wieschollek 2016, Anca Sticlaru 2017];

Materials may have various appearances depending of their surface properties, lighting geometry, viewing geometry, camera settings, etc. It has been shown in few studies, such as [Maxim Maximov 2018, Carlos Vrancken 2019], that combining different views, lighting conditions, etc. may slightly improve the training task. Training an appropriate classifier requires a training set which covers not only all viewing and lighting conditions and capture and processing settings, but also the intra-class variance of the materials. It also requires to annotate all training data that requires a strong effort for large datasets.

In the ideal case, one would like to predict what would be the appearance of a surface whatever the viewing direction and other factors having an impact on the capturing process. It is a quite challenging, ill-posed and under-constrained problem that remains hard to solve for the general case. This can be achieved either using implicit or explicit methods, or using deep learning approaches.
- Implicit methods based on image-based representations enable to interpolate new views from a series of photos taken under a limited set of conditions.
- Explicit methods based on simulation-based representations enable to extrapolate new views from simulation (e.g. changes of lighting geometry, viewing geometry) relating to reflectance model parameters (e.g. Phong).
- Deep-learning based approaches enable to recognize material appearance based on previous knowledge extracted from an image dataset.

Meanwhile classical deep learning approaches extract appearance representations relating reflectance model parameters to a specific illumination model, more recent approaches extract appearance representations over 2D reflectance maps, such as [K. Kim 2017, M. N. Wang TY 2018, Maxim Maximov 2018]. For example, in [Manuel Lagunas 2019] the authors developed a novel image-based material appearance similarity measures derived from a learned feature space (that correlates with perceptual judgements) using a loss function combining physical-dependent appearance features and visual perception-dependent appearance features.

[Grigorios Kalliatakis 2017] investigated whether synthesized data can generalize more than real-world data. They showed that the best performing pre-trained convolutional neural network (CNN) architectures can achieve up to ~91.03% mean average precision when classifying materials in cross-dataset scenarios. They also showed that synthesized data can achieve an improvement on mean average precision when used as training data and in conjunction with pre-trained CNN architectures, which spans from ∼ 5% to ∼ 19% across three widely used material databases of real-world images (FMD, ImageNet7 and MINC-2500).

In Section 3, we will survey the most recent Convolutional Neural Networks (CNN) and Deep Learning architectures which have been proposed in the literature for materials recognition.

**Section 1: Materials databases**

Meanwhile several image databases have been created for object recognition, like ISLVRC 2012 [ISLVRC 2012], significantly fewer databases have been created for material classification. These material databases can be grouped into three categories.

   **1.1 BRTF (bi-directional-reflection-transmittance function) measurements dataset**
Description**:** Images in this kind of datasets are collected under controlled conditions. In order to measure BRTF values of material instances, lighting and viewing conditions are controlled. Advantage**:** BRTF models describe perfectly visual appearance of material instances. Limitations**:** most of research studies based on BRTF measurements focus only on the task of texture or material classification, or on the task of building features that are invariant to pose and illumination. Figure 1 shows an example of two different samples that look quite similar under a diffuse illumination condition, whereas these sample will have a very diverse visual appearance under a directional illumination condition. As a consequence, combining the information about the optical physics properties of the materials, such as in [Manuel Lagunas 2019], should improve the discriminating power of materials recognition methods, or could potentially be used for transfer-learning.

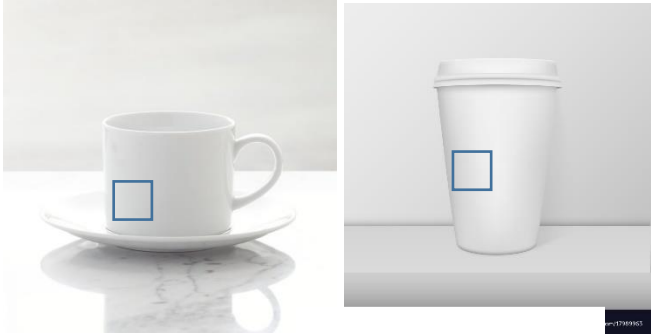

(a) Ceramic coffee cup    (b) Paper coffee cup

Figure 1: Identifying and understanding the interactions between material appearance features (here between light and surface reflectance) may help to categorize image patches.

Representative datasets (see also [Michael Weinmann 2014]):
- Columbia- Utrecht Reflectance and Texture Database (CUReT) [CUReT 1999]: 61 material samples and 205 different lighting and viewing conditions. As only a single material instance is provided per class, no generalization can be done to classifying object categories, due to a lack of intra-class variation.
- KTH-TIPS2 (Textures under varying Illumination, Pose and Scale) database [KTH-TIPS2 2006] which was created to extend the CUReT database by providing variations in scale, as well as pose and illumination, and by imaging other samples of a subset of this database in different settings. As only four samples are provided per category this still limits the representation of the intra-class variance of materials observed in real-world scenarios.
- UBO 2014 [UBO 2014]: 7 material categories (carpet, fabric, felt, leather, stone, wallpaper, wood), each of which contains 12 material instances for being capable to represent the corresponding intra-class variances. 151*151 illumination and viewing directions.

## 1.2 Real-world datasets

Description**:** in contrast to BRTF measurements, here, images are taken in the wild and therefore, material's visual appearance is more various and it depends on material instance, natural light, stochastic pose, etc. Moreover, real-world datasets can be classified relatively to the context information surrounding target materials.

Advantage**:** It becomes possible to exploit one material's invariant element through its various instances.

Challenge**:** most of recent research studies use deep learning network to extract invariant features describing materials. They obtained good results to classifying materials categories. The further step will consist to extract further knowledge from information provided by the networks, i.e. to better analyze/understand unnamed features (see [Gabriel Schwartz 2018]), how to better exploit learned features, how to mind semantic relationship, association patterns, between features (as in [Rachasak Somyanonthanakul 2018]).

Representative datasets (non-public datasets are not listed here, see also [M. Cimpoi 2014, Manuel Lagunas 2019]):
- FMD (Flickr Material Database) [FMD 2014]: 1000 images and 10 categories. A variety of illumination conditions, compositions, colors, texture and material sub-types. This makes this database a very challenging benchmark. Almost no context information.
- DTD (Describable Textures Dataset) [DTD 2014]: 5646 images, 47 classes. No context information. For each image, there is not only key attribute (main category) but also a list of joint "describable texture" attributes [M. Cimpoi 2014].
- GTOS (Ground Terrain in Outdoor Scenes) [GTOS 2017]: 30.000 Images covering 40 common classes in outdoor scenes. No context information.
- MINC (Materials in Context Database) [MINC 2015]: Patches cropped manually from material segments in the wild. With context. A Very big dataset (as shown in Table 1). MINC-2500 is a sub-dataset: in each category there is 2500 images.
- COCO (Common Objects in context) dataset [COSO 2015]: An image semantic segmentation task dataset with 163K images (train 118K, val 5K, test-dev 20K, test-challenge 20K) with annotations for 91 stuff classes and 1 'other' class. It contains several material classes but some of them are classified further into more accurate class, like fabric whose instances appear in 'cloth', 'towel', 'curtain', etc.
- *Open Surfaces* (OS) dataset [S. Bell 2013]. OS comprises 25,357 images, each containing a number of high-quality texture/material segments. Many of these segments are annotated with additional attributes such as the material name, the viewpoint, the BRDF, and the object class. Material classes are highly unbalanced and segment sizes are highly variable.

| Patches | "material" category | Patches | "material" category | Patches | "material" Category related to non-usual material properties |
|---|---|---|---|---|---|
| 564,891 | Wood | 114,085 | Polished stone | 216,368 | Glass |
| 397,982 | Fabric | 188,491 | Metal | 55,364 | Water |
| 14,954 | Wallpaper | 28,108 | Ceramic | Patches | material" category related to semantic features |
| 98,891 | Carpet | 38,975 | Plastic | 465,076 | Painted |
| 29,616 | Stone | 120,957 | Foliage | 64,454 | Brick |
| 83,644 | Leather | 35,246 | Skin | 23,779 | Paper |
|  |  | 142,150 | Sky | 75,084 | Mirror |
|  |  | 26,103 | Hair | 25,498 | Food |
|  |  |  |  | 147,346 | Tile |

Table 1 : MINC patch counts by "material" category. Some categories are quite similar to categories defined for BRTF datasets, such as UBO 2014 (see   ), meanwhile other material categories are missing, such as "felt". Other material categories not covered by most of BRTF datasets are added, such as foliage or sky (see   ). Other challenging material categories are listed, such as "glass" or "water" (two "transparent" material, see   ). Some categories are related to semantic objects, such a "mirror" (which is related to more than one material class) or "food" (a very heterogeneous class) (see   and discussion after).

Other (non-public) datasets:
- "Google Material Database" (GMD)[1] [Patrick Wieschollek 2016]: around 10.000 images divided in 10 material categories and 4868 material samples. The images of this dataset correspond to real-world images captured under various illuminations and views. Once again, as with other databases, the number of images in each GMD category is unbalanced (e.g. 162 images for "paper" up to 1067 for "fabric"), which leads to some classes performing better than others due

---
[1] Not yet publicly available

to this unbalance of images, as shown in Table 2.

| Dataset | FMD | | | | | | ImageNet7 | | | | | |
|---|---|---|---|---|---|---|---|---|---|---|---|---|
| Category | fabric | glass | metal | paper | plastic | wood | fabric | glass | metal | paper | plastic | wood |
| GoogLeNet | 71.07 | 94.03 | 86.3 | 90.57 | 87.25 | 86.87 | 90.41 | 90.4 | 98.28 | 90.29 | 56.67 | 93.47 |
| ResNet50 | 98.9 | 97.59 | 97.31 | 97.15 | 99.65 | 96.98 | 100 | 99.8 | 100 | 99.68 | 99.52 | 100 |
| ResNet101 | 94.56 | 97.15 | 98.52 | 93.88 | 99.04 | 96.8 | 100 | 99.8 | 99.82 | 99.98 | 99.83 | 100 |
| ResNet152 | 98.83 | 96.36 | 97.59 | 99.5 | 99.83 | 99.34 | 100 | 99.6 | 100 | 99.97 | 99.97 | 100 |
| VGG_CNN_S | 75.7 | 97.07 | 82.19 | 85.8 | 82.61 | 86.49 | 86.08 | 87.5 | 97.78 | 89.34 | 46.64 | 92 |
| VGG_CNN_M | 68 | 96.54 | 92.28 | 82.69 | 87 | 85.35 | 80.78 | 83.5 | 94.17 | 82.21 | 44.47 | 91.22 |
| VGG_CNN_F | 55.96 | 95.75 | 81.61 | 83.3 | 88.22 | 85.92 | 78.01 | 86.6 | 94.07 | 80.74 | 46.68 | 88.52 |
| VGGNet16 | 90.92 | 96.68 | 89.66 | 93.21 | 90.14 | 80.79 | 88.69 | 86.5 | 94.28 | 90.86 | 43.35 | 96.48 |
| VGGNet19 | 89.79 | 91.95 | 88.38 | 90.68 | 91.75 | 87.15 | 87.14 | 92.5 | 97.65 | 85.66 | 47.09 | 91.15 |
| Dataset | MINC2500 | | | | | | GMD | | | | | |
| Category | fabric | glass | metal | paper | plastic | wood | fabric | glass | metal | paper | plastic | wood |
| GoogLeNet | 81.39 | 84.97 | 89.51 | 92.87 | 71.04 | 92.52 | 78.17 | 85.5 | 84.04 | 90.23 | 77.61 | 96.91 |
| ResNet50 | 99.97 | 99.95 | 99.98 | 99.99 | 99.81 | 100 | 99.99 | 100 | 99.65 | 99.21 | 98.82 | 99.95 |
| ResNet101 | 99.99 | 99.96 | 99.87 | 99.92 | 99.91 | 99.95 | 99.92 | 99.9 | 99.67 | 99.07 | 99.57 | 99.75 |
| ResNet152 | 100 | 99.9 | 99.98 | 99.96 | 99.99 | 99.95 | 99.71 | 99.8 | 99.41 | 99.73 | 99.23 | 99.91 |
| VGG_CNN_S | 77.21 | 84.76 | 78.07 | 88.66 | 59.81 | 88.84 | 83.87 | 90.9 | 85.89 | 97.07 | 60.76 | 97.19 |
| VGG_CNN_M | 79.85 | 84.76 | 85.51 | 90.15 | 61.39 | 86.47 | 88.52 | 89.5 | 75.34 | 95.86 | 50.01 | 95.95 |
| VGG_CNN_F | 76.85 | 78.84 | 77.16 | 84.67 | 59.97 | 86.05 | 81.37 | 86.4 | 78.72 | 92.29 | 76.18 | 95.04 |
| VGGNet16 | 76.73 | 83.03 | 81.7 | 88.27 | 65.32 | 85.48 | 91.15 | 95.2 | 93.06 | 98.43 | 90.14 | 99.75 |
| VGGNet19 | 81.36 | 70.68 | 77.73 | 90.8 | 61.76 | 79.88 | 88.76 | 92.1 | 88.79 | 97.94 | 86.05 | 98.55 |

Table 2: Accuracy of different networks vs material categories and databases [Anca Sticlaru 2017]. It is not surprising that for some categories (e.g. "fabrics" in GMD) results are not significantly more accurate than some others (e.g. "papers"), even if there is a higher number of data available. This is due to the greater heterogeneity of the class "fabrics". On the other hand, the top three accuracies for GMD are for wood, paper and glass (whatever the network used), these results demonstrate that, when a classifier is able to extract discriminative features from sufficiently distinctive classes, it does need a high number of images.

### 1.3 Synthesized database

Description: material or texture rendering technique is widely used in computer graphics. As material databases are not abundant and synthesized material images can be generated quickly and pixel-wise annotated with no effort, synthesized database seems a good way to enrich existing database.
Advantages: Easy collection and annotation, moreover synthesized datasets can be very large.
Limitation: currently there is no material database which simulates materials in the wild.

Representative datasets:
- Bidirectional Texture Functions (BTF) Material database [Michael Weinmann 2014] based on UBO 2014 [UBO 2014]. At each synthesized image corresponds: a material sample, a virtual illumination condition, and a virtual viewing point. Database size: 84 Material samples (7 categories x 12 samples per category), 30 illumination conditions (6 natural lights x 5 directions), 42 viewing points. So, at each samples correspond 1260 images (30 illuminations x 42 viewing points) and at each category corresponds 15.120 images (1260 images per sample x 12 samples per category). The number of images generated is therefore of 105.840 images (15.120 images per category x 7 categories). This make this dataset very useful when one want to compute material feature invariant to illumination and viewing condition changes. This is very important for some materials classes, such as "metal" or "skin", meanwhile for other less reflective materials, such as "carpet", this is less effective.
- "Playing for data Ground Truth from Computer Games" [Stephan R. Richter 2016]: Like the COCO dataset, this dataset is dedicated for sematic segmentation and contains many stuff segments.

The source is a video game: "Grand Theft Auto 5" where a virtual world is created in a way to imitate the real world's scenes. It cannot be directly used as a material database, but it is a good start to know how to collect synthesized material segments in a virtual "wild".

Other (non-public) dataset:
- The Material Appearance Database[2] introduced in [Manuel Lacunas 2019], which contains 9,000 rendered images (using the Mitsuba physically-based renderer), depicting objects with varying materials (100 different materials, such as paints, metals, fabrics, or organic materials), shapes (15 different 3D models, with varying complexity, curvature, convexity, such as sphere, blob, bunny shape) and illuminations (6 illumination conditions, such as outdoor illumination environment). It also includes measured BRDFs (from the MERL BRDF database) with reflectance data captured from real materials, surface normals, depth, transparency, and ambient occlusion maps.

Optical models, such as BRTF, can perfectly describe material instance but they contribute few to understanding material classes and material features. Therefore, in recent years, researchers started to collect databases in the wild and to recognize materials in class-level, not instance-level anymore.

It would be interesting to study if and how context influences material recognition because some existing databases are diverse at the scale of context. It would be also interesting to study how synthesized material dataset could be used to enrich current real-world datasets, as generating synthesized datasets is easier compared to real images. However, to the best of our knowledge, currently there is no yet relevant synthesized material dataset in the wild. May be collection from video game could be a good choice.

In the ideal case, to overcome the problems of existing material databases which only include a limited number of classes and of viewing conditions, one would like also to use a learning model (e.g. a learning to learn model) that would be able to predict the appearance of any material in new views or capture conditions, and generalize well even if images are acquired under complex real-world scenarios. This requires the ability to refine in a dynamic way the learned model with new observations.

**Section 2: Influence of the context**

Context plays a very important role in material recognition. Local information, such as texture and color, is not sufficient for recognizing material categories but they may be useful in some study cases. For example, meanwhile color is an efficient feature to recognize "skin" or "sky" materials, color is an irrelevant feature to recognize "carpet" materials, considering the intra-class variance of this class. In other study cases, contextualization is essential. For example, in Figure 2, if we just crop a local patch from a coffee cup, the patch appearance suggests a smooth white material, so maybe it belongs to the "ceramic" class or to the "paper" class. But, if we scale up the view until we see the surrounding context, this time, anyone will be pretty sure that the patch (surrounded in blue on the left image) corresponds to a "ceramic" due to the shape and the reflectance properties of the object to which it belongs, meanwhile for the patch (surrounded in blue on the right image) we could state that this patch belongs to the "paper" category. On the other hand, if we crop another local patch on the table (e.g. the patch surrounded in orange on the left image), the patch appearance suggests a smooth white material similar to the first selected patch, so maybe it also belongs to the "ceramic" class. But, if we scale up the view until we see the surrounding context, this time, anyone will be pretty sure that this patch corresponds to a "stone" class (sub-class "marble") due to the texture and the reflectance properties of the surface to which it belongs.

---
[2] Not yet publicly available, see http://webdiis.unizar.es/~mlagunas/publication/material-similarity/#downloads

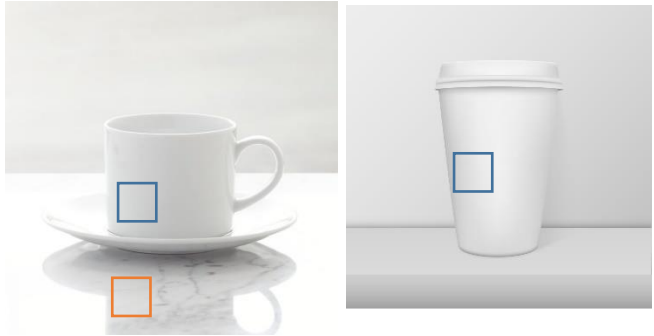

(a) Ceramic coffee cup    (b) Paper coffee cup

Figure 2: The contextualization enables to extract more relevant features than the ones extracted from local samples (independently of the context, such as shown in small windows).

Few articles support and prove this theory (e.g. [R. Mottaghi 2014, Sean Bell 2015, G. Lin 2017]), as discussed below.

In [Sean Bell 2015] the authors studied the influence of the context on material recognition. After collecting material segment and determining the center of patch, in order to crop fixed- sized patch for training a CNN classifier, a patch scale method was implemented to decide how much context is included into the patch. Thus, when the scale goes up, more context is present in the patch and vice versa. For example, in the Figure 3 below, the size of the patch is fixed but when the scale increases, the context of the target patch (here the coffee cup) becomes more obvious. In order to determine the patch scale, the authors measure classification accuracies from several scales, as shown in Figure 3 (a). The optimum patch scale found was of 23,3% of overlap of the scene area by the patch. The authors also performed a class-wise test in order to study the influence of material categories on optimum patch scale. Figure 3 (b) illustrates that almost all the classes can be accurately recognized if their patches contain appropriate quantity of context. In some cases, such as for the "sky" class, the context contributes to slightly improve the mean class accuracy (from 92,5 % to 98,3%), meanwhile for more complex cases, such as the "mirror" class, the mean class accuracy significantly increases, from 45,7% to 85.6%.

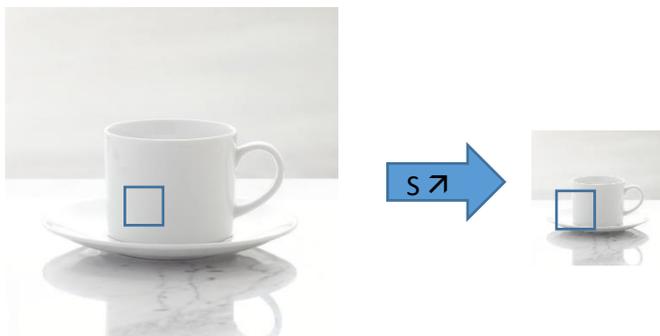

Figure 3: Illustration of the patch scale method.

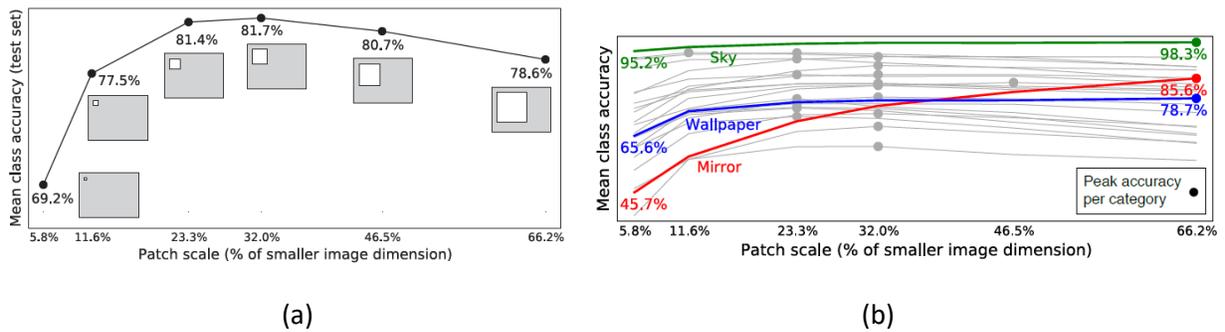

(a) (b)

Figure 3: (a) computation of the optimum of the patch scale, (b) accuracy vs patch scale per category [Sean Bell 2015].

In Figure 4, we can see from the cropped patches (at a chosen scale) the influence of the context on material categorization. For some material categories, such as "foliage", the patch text accuracy is very good, meanwhile for other categories, such as "polished stone", the patch accuracy is rather low due to the intra-class variance of these classes and the inter-classes variance between some classes (e.g. between "polished stone" and "water"). This means that if the context is a very important parameter to take into account to classify some types of materials, it is not the only key parameter. In some cases, the ambiguity that exists between two classes cannot be solved using the context only. Another question arise here, when we look at the class "polished stone", does that make sense to categorize all samples of this class in only one class ? Considering the intra-class variance of this class may be that will make sense to sub-categorize this class in several sub-classes, but will that contribute to improve the confidence prediction?

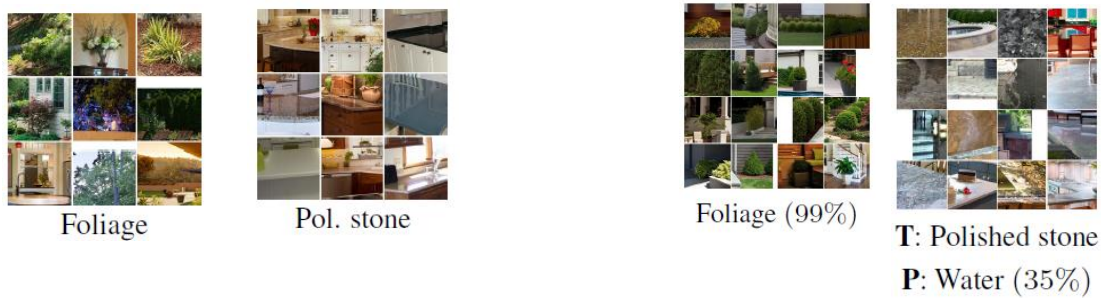

Figure 4:
(a) Examples of patches from the MINC database sampled so that the patch center belongs to the category in question (which is not necessarily the case for the entire patch).

(b) confidence prediction using GoogLeNet CNN

From the former discussion it appears that several material categories in existing datasets have not been carefully defined. To address this issue, Schwartz et al. proposed in [Gabriel Schwartz 2018] a set of material categories derived from a materials science taxonomy. Additionally, they created a hierarchy based on the generality of each material family. Their hierarchy consists of a set of three-level material trees. The highest level corresponds to major structural differences between materials in the category (such as metals, ceramics, polymers). The mid-level categories as groups that separate materials based primarily on their visual properties (such as matte appearance, color variations). The lowest level, fine-grained categories, can often only be distinguished via a combination of optical physics and visual properties (such as reflectance, transmittance). Meanwhile this hierarchy seems to be sufficient to cover most natural and manmade materials, certain "mid-level" materials categories (such as food, water, and non-water liquids) do not fit within the strict definitions described above. It will be interesting to investigate the effectiveness of the taxonomy proposed by [Gabriel Schwartz 2018] in terms of intra-class variance reduction / inter-class variance increase (as example see Figure 5), and confidence prediction.

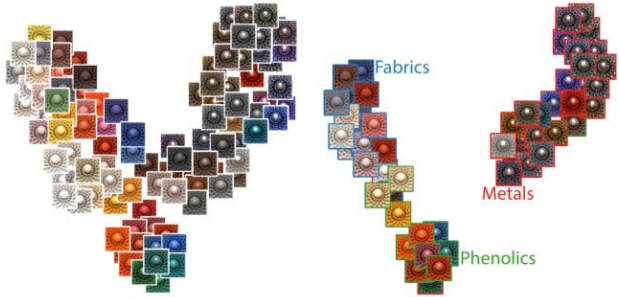

Figure 5: Visualization of the MERL dataset in a 2D space (reflectance vs color) based on the feature vectors provided by [Manuel Lagunas 2019]. Left: All images of the MERL dataset (from left to right the reflectance significantly increases, meanwhile color variations from bottom to up are less significant). Right: Materials from three different categories only ("metals", "fabrics", and "phenolics"). Here the intra-class variance of "fabrics" is greater than the inter-class between "fabrics" and "phenolics", as a consequence it is more challenging to cluster these two material categories than the "metals" category, from their material appearance similarity using these 2 dimensions only.

Unlike to [Sean Bell 2015], where front ground material and its context information together go into the classifier, in [G. Schwartz 2016] the authors proposed another method which first extracts separately features from local material patch and from its context (objects and scenes). Next, the method concatenates these feature maps all together and material prediction is made based on this fused feature maps. The purpose of this method is to achieve dense prediction which is more difficult than image classification. Table 3 below demonstrates that both objects and scenes (context parameters) have a strong effect on materials recognition. It also shows that the combination of objects and scene provides significantly higher confidence prediction than either of the two alone, suggesting that knowing both objects and scenes provides unique cues not present in either individual category group.

| Context | Accuracy | Mean class accuracy |
|---|---|---|
| None | 63,4% | 60,2% |
| Only Scenes | 68,2% | 67,7% |
| Only Objects | 67,0% | 63,5% |
| Scenes + Objects | 73,0% | 72,5% |

Table 3: per-pixel average accuracy with no context, as well as with each separate form of additional context. Results obtained from diverse materials database, such as COCO [G. Schwartz 2016].

One way to evaluate the impact of the context is to evaluate the change of recognition accuracy with or without context. We compared the recognition accuracy of patches taking into account the background context and of patches after removing the background context. We used the FMD dataset [FMD 2014] as it is very easy to produce images without background context using a mask covering material region (as illustrated in Figure 6).

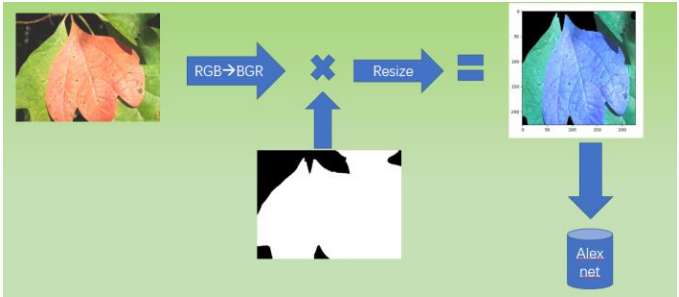

Figure 6: masking process to remove the background context and to resize the resulting image to the original image size.

We used the following process:

1. We denote the training dataset containing original images as trainC and the training dataset containing images without context as trainNC;
2. Identically, we denote the test dataset containing original images as testC and the test dataset containing images without context as testNC;
3. We remove in both test datasets images which do not contain context in the first place;
4. We trained and tested two identical AlexNet pre-trained on ImageNet, respectively on {trainNC, testNC} and {trainC, testC};

From this very simple test, we demonstrated that the background context contributes in some way to materials classification, as shown in Table 4.

| Test / Train | TestC | TestNC |
|---|---|---|
| TrainC | $79.5 \pm 3.32$ | |
| TrainNC | | $77.5 \pm 3.32$ |

Table 4: Number of wrongly classified images with or without taking into account the background context.

### Section 3: Deep learning for material recognition

Since several years Convolutional Neural Networks (CNN) and deep learning architectures have proved their efficiency in the domain of object recognition. Since few years only few deep learning architectures have been investigated in the domain of material recognition. In the following, we will survey the main methods recently published.

*Transfer Learning: a promising approach*

In [Patrick Wieschollek 2016] the authors exploited the potential of transfer learning for material classification. As it is not possible to train a deep learning convnet on a small dataset, such as FMD , the authors proposed to transfer the structure and weights of a convnet (CNN) trained for object classification and augment it to boost the performance of material classification. The training of the convnet is done using a pre-trained model (AlexNet) with fewer classes and more data per category from the ImageNet ILSVRC 2012 challenge. The structure of the convolutional layers (acting as local filters) and fine-tuned layers of this convnet is illustrated in Figure 7. By limiting the amount of information extracted from the layer before the last fully connected layer, transfer learning was used to analyze the contribution of shading information, reflectance and color to identify the main characteristics which determine into which material category an image belongs to. According to A. Sticlaru [Anca Sticlaru 2017] it would be also interesting to access the information from the last convolutional layer, as it could provide information about the texture of the material, and to see how the results and the overall system would be affected.

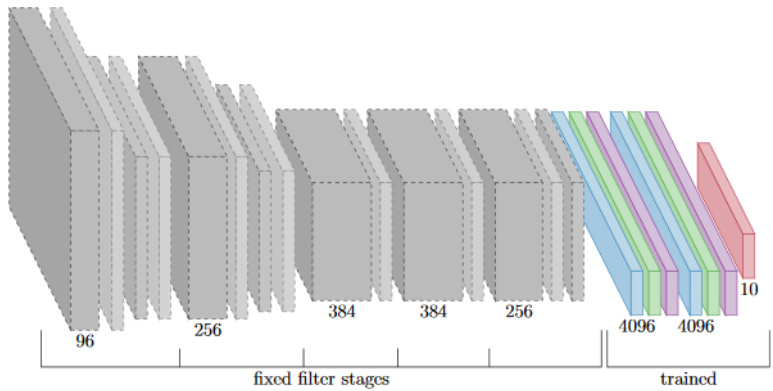

Figure 7 : architecture of the Deep convnet used for transfer learning [Patrick Wieschollek 2016]

The performance of the convnet was tested on FMD dataset. Results reported in [Patrick Wieschollek 2016] clearly demonstrated the potential of the method proposed. They also shown that object and material recognition have some similar invariants when using deep learning mechanism, and that object's features extracted from ImageNet are transferable to material's features.

The authors also tested various versions of the described convnet layout on the Google Material Database (GMD) by fixing different combination of filter stages. As for the FMD dadaset, the parameters of the layers were initialized by the pretrained model from the ImageNet ILSVRC 2012 challenge. The convolutional layers set during training on the FMD were kept fixed, meanwhile the fine-tuned layers were retrained on the GMD (which contains 10 times more images than FMD). The performance of the resulting deep convnet significantly increased on material classification in natural images. The accuracy is of 74% with GMD, meanwhile the accuracy with FMD is only of 64%. This confirms the significant influence of the dataset used on the performance of the training. As demonstrated in [Anca Sticlaru 2017], the best performance is achieved form largest real-world dataset (i.e. having the most images) when state-of-the-art datasets are used for training and testing.

The main conclusion one can draw from [Patrick Wieschollek 2016] is that object cues, such as shape and reflectance, are beneficial to material recognition, even probably essential.

*Local orderless representation vs global representation*

In [M. Cimpoi 2015] the authors proposed a new texture descriptor, named FV-CNN, built on Fisher Vector (FV) pooling of a Convolutional Neural Network (CNN) local filter bank. FV pools local features densely within the described regions removing global spatial information, and is therefore more apt at describing materials than objects. Orderless pooled convolutional features are extracted immediately after the last linear filtering operator and are not otherwise normalized. FV is computed on the output of a single (last) convolutional layer of the CNN. The dense convolutional features are extracted at multiple scales and orderless pooled into a single FV, thus by avoiding the computation of the fully connected layers, the input image does not need to be rescaled to a specific size. As a consequence, FV-CNN can seamlessly incorporate multiscale information and describe regions of arbitrary shapes and sizes. The structure of the FV-CNN is illustrated in Figure 8.

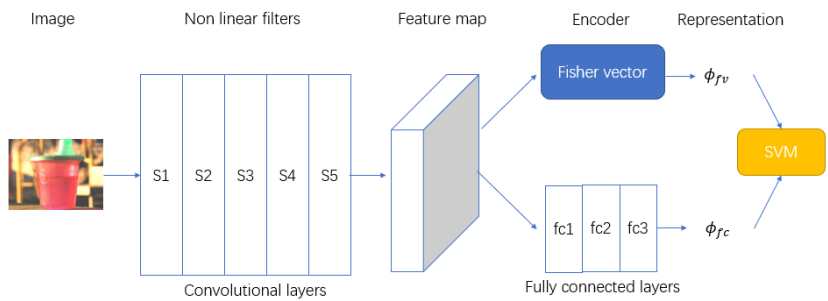

Figure 8 : architecture of the FV-CNN. Firstly, non-linear filter banks are used to extract features, then the output feature map is pooled and encoded as Fisher vector. This encoding process treats independently each feature vector of the feature map. The spatial relations between feature vectors are not taken into account. So, the Fisher vector encoded is an orderless representation of the input material image. Lastly, a SVM classifies the input image with the representation.

M. Cimpoi et al. demonstrated that FV-CNN easily transfers across domains without requiring feature adaptation as for methods that build on the fully-connected layers of CNNs. They trained their CNN network on ImageNet's ILSVCR, and then, orderless filter banks were retrained on the FMD and on the MIT datasets. The FV-CNN achieved 79.8% accuracy on FMD and 81% accuracy on MIT, providing absolute gains of more than 10% over existing approaches.

Results shown in Table 5 demonstrate that FV-CNN outperforms fine-tuned CNN, such as FC-CNN and that the combination of FC-CNN and FV-CNN increase the classification accuracy.

|  | VGG-19 | | |
|---|---|---|---|
|  | FV-CNN | FC-CNN | FV+FC - CNN |
| FMD dataset | 79.8 $\pm$ 1.8% | 77.4 $\pm$ 1.8% | 82.4 $\pm$ 1.5% |

Table 5: accuracy of FV-CNN in comparison to FC-CNN [M. Cimpoi 2015].

The main conclusions one can draw from [M. Cimpoi 2015] is that orderless pooling of CNN features is very beneficial to material recognition and that object context's information (features of FC-CNN) helps material classification.

Creating an orderless representation becomes a mainstream in material recognition world. Although the FV-CNN [M. Cimpoi 2015] showed excellent performance, the main problem with FV-CNN is that feature extraction, encoding of Fisher vector, and SVM classifier training are self-contained. As a result, classification loss cannot be back-propagated into the whole structure. In other words, the entire pipeline is not structured in an end-to-end learning manner. To solve this problem, [H. Zhang 2017] proposed Deep Ten, another network based on a texture encoding layer which acts like CNN's fundamental layers (unlike Fisher vector encoding which is just a stand-alone encoding process) which are capable of feed-forward and back-propagation. Furthermore, Deep Ten inherits the legacy of FV-CNN and also encodes feature vectors as orderless representation.

Results shown in Table 6 demonstrate that Deep Ten outperforms FV-CNN on 4 datasets:

| Database<br>Network | MINC 2500 | FMD | KTH-TIPS2 | GTOS |
|---|---|---|---|---|
| *FV-CNN* | *61.8* | $79.8 \pm 1.8\%$ | $81.8 \pm 2.5\%$ | 77.1 |
| *Deep Ten* | **80.6** | $\mathbf{80.2 \pm 0.9\%}$ | $\mathbf{82.0 \pm 0.9\%}$ | $\mathbf{82.0 \pm 3.3}$ |

Table 6: accuracy of Deep Ten in comparison to FV-CNN on 4 datasets. Deep Ten is better than FV-CNN, especially on MINC 2500. This is due to the Deep Ten's end-to-end learning manner which allows to optimize CNN features when a dataset is large-scale, such as MINC-2500. Such optimization cannot happen with FV-CNN because its CNN features cannot be updated as they are obtained by fixed pre-trained CNN.

As the combination of local orderless representation of FV-CNN and global representation of FC-CNN boosts classification accuracy further, is it possible to equally improve the recognition when combining Deep Ten's encoding layer's output with global representation of FC-CNN? To achieve this objective Jia Xue et al. proposed in [Jia Xue 2018] the Deep Encoding Pooling Network (DEP). The structure of this network is illustrated in Figure 9.

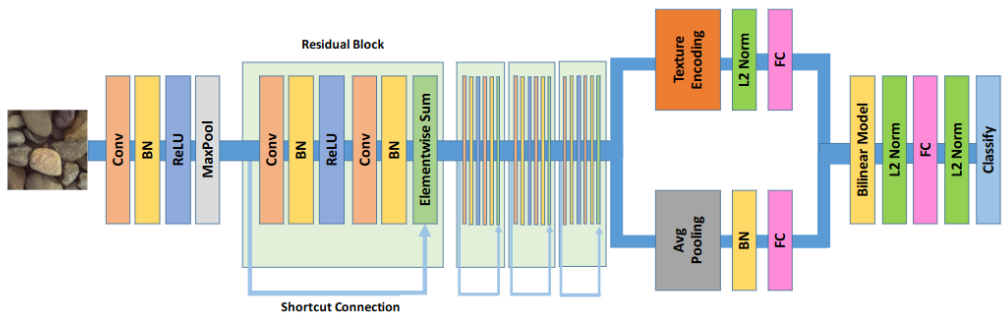

Figure 9: architecture of the DEP [Jia Xue 2018]. Outputs from convolutional layers are fed jointly into the encoding layer and global average pooling layer. The outputs of these two layers are processed with bilinear model. Such structure perfectly combines local and global representation together, meanwhile the end-to-end learning is preserved.

Results shown in Table 8 demonstrate that DEP slightly outperforms Deep Ten and FV-CNN on 2 datasets:

| Network \ Database | DTD | MINC 2500 |
|---|---|---|
| FV-CNN | 72,3% | 63,1% |
| Deep Ten | 69,6% | 80.4% |
| **DEP** | **73,2%** | **82,0%** |

Table 8: accuracy of DEP in comparison to Deep Ten and FV-CNN [Jia Xue 2018].

As in [H. Zhang 2017], DEP employs convolutional layers with non-linear layers from ImageNet pre-trained CNNs as feature extractors. Transfer learning from object recognition to material recognition is performed using global features. It could be also performed using object context's features. [Jia Xue 2018] show that the combination of global features with local features provides more discriminative information to their classifier and thus increases its performance. Likewise Deep Ten is based on the same idea. Their main contribution to the state-of-the-art is to make the whole process structured in an end-to-end learning manner, to the benefit of the classification, especially for large-scale dataset.

Inspired by the former global-local structure, we designed a multi-task network. The structure of this network is illustrated in Figure 10. This network uses two conv3*3 which are added to VGG-19 convolutional layers. This network is structured in two branches. One of them uses a 1*1 convolutional layer, as local classifier, to recognize materials according to each feature vector of the feature map. This branch is only activated during the training of the local classifier, meanwhile during the inference only the main branch is activated. Thanks to this additional branch, local feature vectors computed from feature maps contain more semantic information provided from their respective local receptive field. In our tests we used the FMD dataset, as in this dataset local receptive fields have been labelled by masks. Moreover, the two 3*3 convolutional layers learn more efficiently in a supervision manner. As a result, this network produces more performant feature maps than without this branch.

The training procedure is divided in two steps. Firstly, it trains the two 3*3 convolutional layers and the 1*1 convolutional layer of the local classifier branch. Next, it fixes the two 3*3 convolutional layers and trains the fully connected layers of the main pipeline.

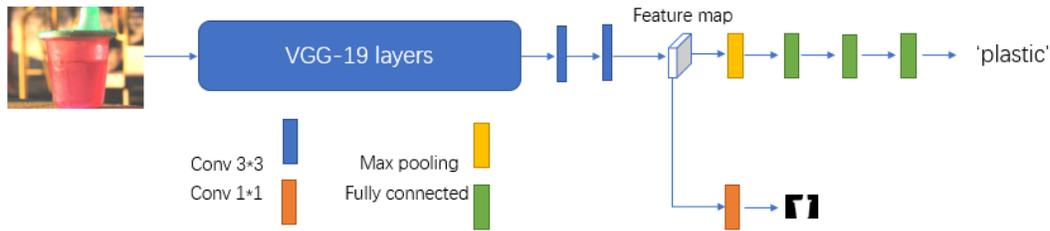

Figure 10: architecture of the multi-task network proposed. The main pipeline corresponds to the upper branch meanwhile the lower branch is devoted to the training of the local classifier.

In order to demonstrate the added value of this additional branch to the network, we also implemented a simplified version of this network (without the additional branch) using the main pipeline described above as a baseline. We trained this simplified network as follows: firstly fixing the two 3*3 convolutional layers, next training the fully connected layers in the main pipeline.

Results shown in Table 8 demonstrate that the multi-task network performs well for large-scale dataset.

| Network  Dataset | baseline | multi-task network |
|---|---|---|
| FMD with only 400 training samples | 76.3 $\pm$ 1.18% | 76.2 $\pm$ 1.75% |
| FMD with 800 training samples | 78.4 $\pm$ 2.43% | 78.9 $\pm$ 2.02% |

Table 8: accuracy of the multi-task network. The performance of the network (78.9 %) is lower than the FV-CNN (79,8 %) and Deep Ten (80,6 %, see Table 6). This is not surprising as the number of training samples used to train the two 3*3 convolutional layers is not sufficient to train accurately this multi-task network. On the other hand, with the baseline these layers are frozen. The performance of the network increases when the number of training data grows. Actually, both methods face overfitting problems due to the insufficient number of training data in FMD, better results could be obtained using a large-scale dataset, such as MINC.

*Data augmentation from synthesized dataset: another promising approach*

To the best of our knowledge, only two papers [Michael Weinmann 2014, Grigorios Kalliatakis 2017] proposed to use synthesized material images for data augmentation in the context of material recognition. Both proposed to use the same synthesized dataset as training dataset and to apply trained classifier to real-world test images. The results provided demonstrate the potential of this kind of approach.

In [Michael Weinmann 2014] the authors used the BTF database as material images synthesized simulate real-world environment, as illustrated in Figure 11. Then, based on features extracted from these synthesized images, a SVM classifier is trained and used to infer class of real images. The paper does not show any real-world images used as test database but according to the authors a manual segmentation was performed on each image, this means that test images correspond to masked material regions (i.e. they do not contain any background or context information).

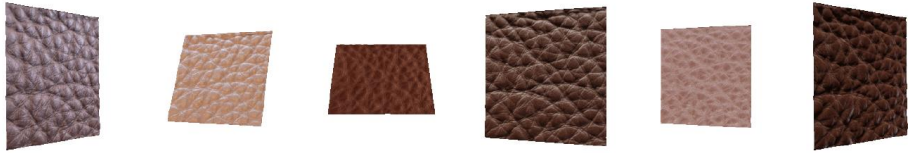

Figure 11: Examples of synthesized images of the same material instance demonstrating the large variation of the BTF database under different viewing and illumination conditions [Michael Weinmann 2014].

The results provided in [Michael Weinmann 2014] demonstrate that the training from internet + synthesized images work better than the training from internet images only (the accuracy falls from 66,67 % to 41,9 % with internet images only). This means that some discriminative information hidden in real-world images have been extracted from synthesized images. They also demonstrate that when the number of synthesized images increases, the performance of the classifier significantly improves (the accuracy increases from 66,67 % to 72,38 % when the number of images is multiply by 4).

Meanwhile the paper [Michael Weinmann 2014] demonstrates that synthesized images can really enrich real-world dataset, it did not address the following questions. First, what is the main advantage to train a deep network on synthesized images? To train a performant deep learning network, as discussed in previous sections, a large-scale dataset is necessary, consequently synthesized dataset can be seen as an efficient choice in case of lack of real-world images. Unfortunately, in the paper deep learning methods are not investigated so this question remains to be answered. Second, do the results provided by [Michael Weinmann 2014] could be reproduced? We do not know what real-world images were used in this study. Maybe tests on more common material datasets, such as FMD, MINC-2500 could be more convincing. These questions have been addressed in [Grigorios Kalliatakis 2017].

In [Grigorios Kalliatakis 2017] three datasets were used: FMD dataset, MINC-2500 and ImageNet 7. All images in these datasets were taken in the wild. Three CNNs with different depths (CNN-F, CNN-M and CNN-M) were used. As baselines, each CNN was trained and tested using each of these three datasets. Next, in order to demonstrate that synthesized images significantly contribute to improve material recognition another test was performed using synthesized dataset as training dataset and real-world datasets as testing datasets.

The results provided in [Grigorios Kalliatakis 2017] demonstrate that once the networks are trained on synthesized dataset, they perform better on every test dataset, compared to the same structure networks trained on the 3 real-world datasets. Moreover features learned on a synthesized dataset generalize well among the three real-world datasets. It would be interesting to reproduce the tests above with more complex real-world scenarios to better understand what these features represent, what are the most efficient features to describe the visual appearance of materials.

**Conclusion**

In this paper, we have reviewed the recent approaches proposed in the literature around material classification. In the first section, we have presented a list of the most widely used datasets in this area, staring from the datasets acquired under controlled conditions to the synthesized datasets, passing through the real world datasets. Along this paper, we have shown that all these datasets provide different but complementary information that could be exploited by any deep learning solution.

Then, we have studied the importance of the contextual information in order to characterize materials. Some papers have shown that this information is essential to improve the classification results, and even that the material and context features should be extracted independently is possible. Our own tests on the FMD dataset also prove the importance of the contextual information for this task.

Finally, we have presented the main deep learning solutions and underlined that the local orderless approaches are the recent trends and provide very good results. In this section, we have also shown that transfer learning helps to learn deep neural networks on small datasets and that synthesized data can be exploited as data augmentation.

In the area of material classification, man works have still to be done. Indeed, we notice several important questions without clear answers yet. For example, knowing the intra-class diversities of the material datasets, should we use hierarchical classes or fine-grain classification ? Or what is the best way to combine local orderless features with global ones ?